\documentclass[letterpaper, 10 pt, conference]{ieeeconf}
\IEEEoverridecommandlockouts

\usepackage{amsmath,amsfonts}
\usepackage{algorithmic}
\usepackage{algorithm}
\usepackage{array}
\usepackage[caption=false,font=normalsize,labelfont=sf,textfont=sf]{subfig}
\usepackage{textcomp}
\usepackage{stfloats}
\usepackage{url}
\usepackage{verbatim}
\usepackage{graphicx}
\usepackage{cite}
\usepackage{siunitx}
\usepackage{multirow}
\usepackage{color}

\begin{document}

\title{Shape Sensing of Continuum Robots using Direct Laser Writing}

\author{Amber K. Rothe*, \IEEEmembership{Graduate Student Member, IEEE,} Nidhi Malhotra, \IEEEmembership{Graduate Student Member, IEEE,} \\and Jaydev P. Desai, \IEEEmembership{Fellow, IEEE}%
\thanks{Research reported in this publication was supported in part by Developmental Funds from the Winship Cancer Institute of Emory University. } 
\thanks{The authors are with the Medical Robotics and Automation (RoboMed) Laboratory, Wallace H. Coulter Department of Biomedical Engineering, Georgia Institute of Technology, Atlanta, GA 30332 USA
        (*A. K. Rothe is the corresponding author. Email: {\tt\footnotesize amber.rothe@gatech.edu}).}%
}



\maketitle

\begin{abstract} 
Continuum robots offer a promising approach for minimally invasive and natural-orifice surgical procedures due to their inherent compliance and dexterity. However, this flexibility also makes estimating the current shape of the robot challenging. Several approaches have been used to reconstruct the shape of these robots, including imaging, optical sensing, magnetic sensing, and resistive sensing. Strain sensors fabricated using direct laser writing (DLW) could provide an alternative sensing method. This technique involves using a laser to induce carbonization of certain polymers to create graphene patterns, such as strain sensors. In this paper, we demonstrate how a flexible continuum joint and a DLW sensor can be machined as one monolithic structure using the same laser and the same setup. The fabricated sensors are characterized using linear and nonlinear models, which are used to predict the joint angle with error as low as \SI{1.76}{\degree}. Furthermore, we demonstrate how a DLW sensor can be used to implement closed-loop control in a robotic joint, achieving tracking error under \SI{3}{\degree}. 

\end{abstract}



\section{Introduction}

 
Continuum robots are a promising approach for minimally invasive surgery due to their flexibility and dexterity. However, their inherent compliance makes pose estimation challenging, and it remains an active area of research, where many approaches have been proposed \cite{Shi2017ShapeSurvey, Sincak2024SensingReview}. Imaging-based methods generally have been used successfully for shape feedback \cite{Shi2017ShapeSurvey}; however, due to the slow scan speed of magnetic resonance imaging \cite{Erin2020MagneticRobotics}, low resolution and high noise of ultrasound \cite{Shi2017ShapeSurvey}, and radiation risk associated with fluoroscopy \cite{Shi2017ShapeSurvey}, safely and reliably achieving a sampling frequency sufficient for closed-loop control with imaging methods is challenging. Another common approach for pose estimation relies on electromagnetic sensors \cite{Zhang2017ShapeTracking, Wang2017PilotAlgorithm}. However, other electronics in the operating room can interfere with these sensors \cite{Sincak2024SensingReview, Shi2017ShapeSurvey}. Furthermore, optical sensors such as fiber Bragg gratings have been used to detect the strain in continuum robots \cite{Park2010Real-TimeInterventions, Sheng2019ARobots}, but they are fragile and expensive. Resistive sensors, which transduce bending strain as resistance change, can take various forms \cite{Sincak2024SensingReview}. Some works incorporate off-the-shelf strain sensors in continuum robotic actuators to obtain shape feedback \cite{Gerboni2017FeedbackSensors, Zhao2022ShapePayloads}. 
 Other works incorporate carbon-filled polymer coatings to create strain sensors \cite{Chen2013Two-axisEndoscope, Pan2025ShapeSensors}. 
Another approach involves encapsulating conductive liquid, such as gallium-indium ink, in a narrow channel \cite{Moyer2025FabricationRobots, Alatorre2022ContinuumApproach}. 
However, the use of resistive strain sensors for continuum robot shape sensing involves several challenges. Many fabrication approaches require multiple fabrication and assembly steps to create the sensor itself and integrate it with the robot \cite{Malhotra2024DesignIdentification}, resulting in high costs and a high likelihood of errors during the assembly procedures. Additionally, miniaturizing the strain sensors sufficiently to allow integration with slender continuum robots such as catheters and guidewires is an open problem \cite{Pandya2016TowardsMeasurement}.

One promising method for fabricating strain sensors is direct laser writing (DLW), wherein a laser is used to create patterns of conductive graphene on certain polymer materials such as polyimide \cite{Srinivasan1994UltravioletNetwork, Lin2014Laser-inducedPolymers, Biswas2023FemtosecondSensor, Duan2019DirectTubes}. 
In prior work, DLW was used to fabricate strain sensors on commercial polyimide tape and which were incorporated with nitinol-based continuum robotic joints \cite{Rothe2025TowardsSensors}. The approach allowed both the DLW sensors and the notched tube continuum robotic joints to be manufactured using the same laser micromachining system. The fabrication did not use expensive materials or additional equipment. However, this approach still required assembling the sensor on the robotic tool body, which was time-consuming and provided an opportunity for errors. In another work, it was shown that tubes made of polyimide can be used to fabricate tendon-driven notched-tube style robotic joints \cite{Malhotra2025DesignApplications}. In this work, we combine these approaches, fabricating DLW strain sensors directly on the robot body, then cutting the notches on the same machine without changing the setup. This approach streamlines the fabrication process and eliminates assembly errors due to misalignment of the sensor. In addition, the present work characterizes the DLW sensors more thoroughly, investigating effects such as hysteresis, rate dependence, and relaxation which were not considered in \cite{Rothe2025TowardsSensors}.
To our knowledge, this is the first attempt to fabricate a DLW sensor directly on a robotic tool.
In this paper, we present the following contributions:
\begin{itemize}
\item{A novel process for fabricating DLW strain sensors on the outside surface notched-tube continuum robotic joints using one machining setup and leaving the central lumen open.} 
\item{A detailed characterization of the proposed sensor-integrated robot (sensing joint) investigating hysteresis, sensor response rate-dependence, and joint relaxation.} 
\item{An assessment of the proposed sensors' accuracy for estimating the bending angle of the sensing joints, using joints with one or multiple sensors.}
\item{A closed-loop control scheme incorporating the sensor feedback from the proposed approach to track the bending angle of the robotic joint.}
\end{itemize}

This paper is organized as follows: Section\,\ref{section_methods} outlines the design 
and fabrication of the sensing joints. Section\,\ref{section_modeling} describes the models used to characterize the sensor.
Section\,\ref{section_experiments} details the experiments, results, and discussion pertaining to the characterization and validation of the sensors. 
Conclusions and future work are discussed in Section\,\ref{section_conclusion}.

\section{Methods}\label{section_methods}
\subsection{Design}\label{section_design}
\begin{figure}[tb]
\centering
\includegraphics[width=0.8\columnwidth]{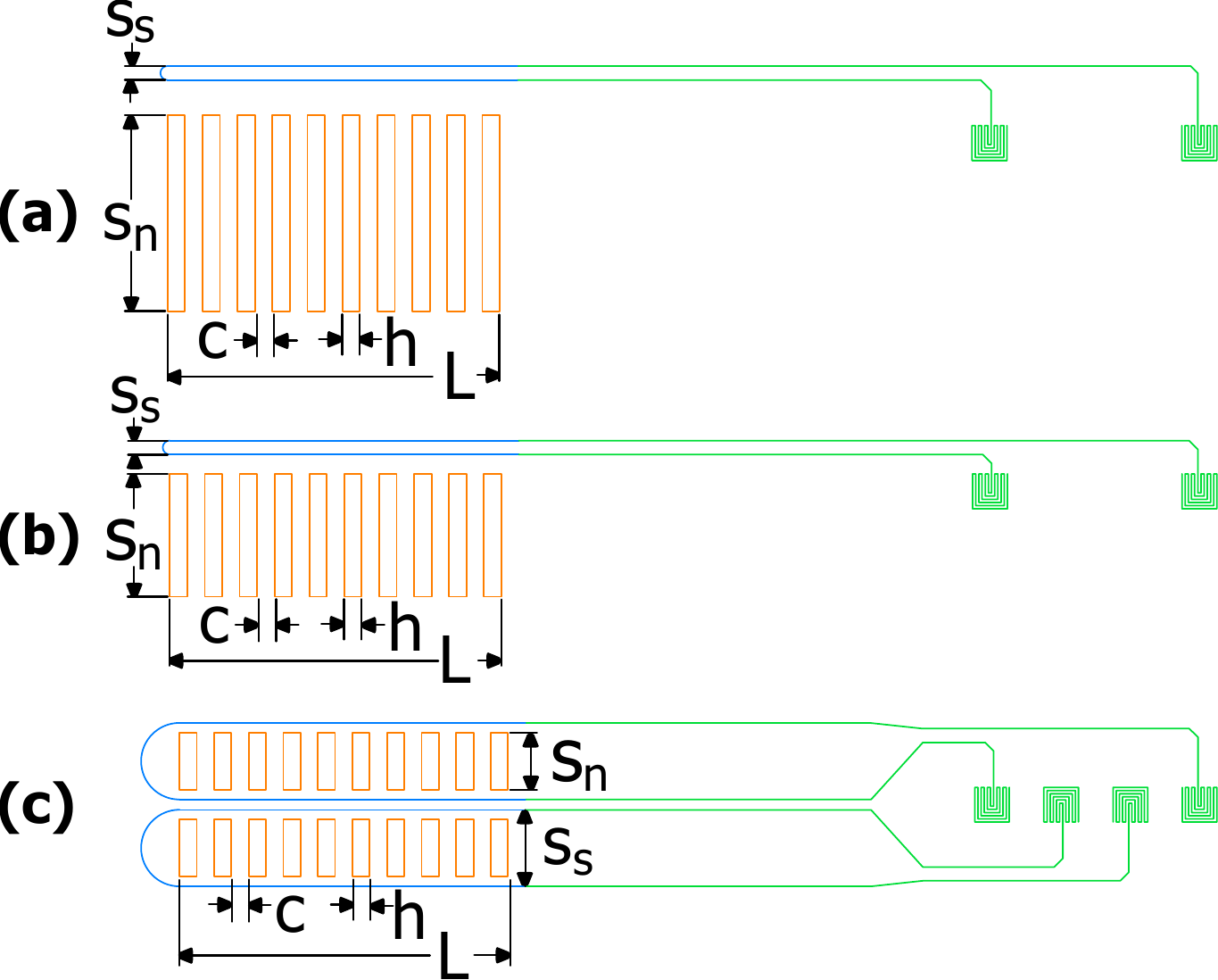}
\caption{Schematic showing the design for (a) Joint A, (b) Joint B, and (c) Joint C. The notches are shown in orange, the strain gauge in blue, and the leads and contact pads in green. }
\label{fig_cad}
\end{figure}

Three sensing joints were designed to be fabricated from stock polyimide tubing (MicroLumen, Oldsmar, FL, USA). Two were unidirectional asymmetric notch (UAN) joints with a single sensor along the unnotched backbone. The other joint was a bidirectional symmetric notch (BSN) joint with two sensors between the notches. The sensors terminated in square spirals to provide large contact pads for connecting additional electronics. Both the sensing element and notches for each joint were designed in the same AutoCAD file (Autodesk, San Francisco, CA, USA). The dimensions of Joint A, Joint B, and Joint C are given in Table\,\ref{table_tube_params} and shown in Fig.\,\ref{fig_cad}(a), Fig.\,\ref{fig_cad}(b), and Fig.\,\ref{fig_cad}(c), respectively. 

\subsection{Fabrication}\label{section_fabrication}

\begin{figure}[tb]
\centering
\includegraphics[width=0.80\columnwidth]{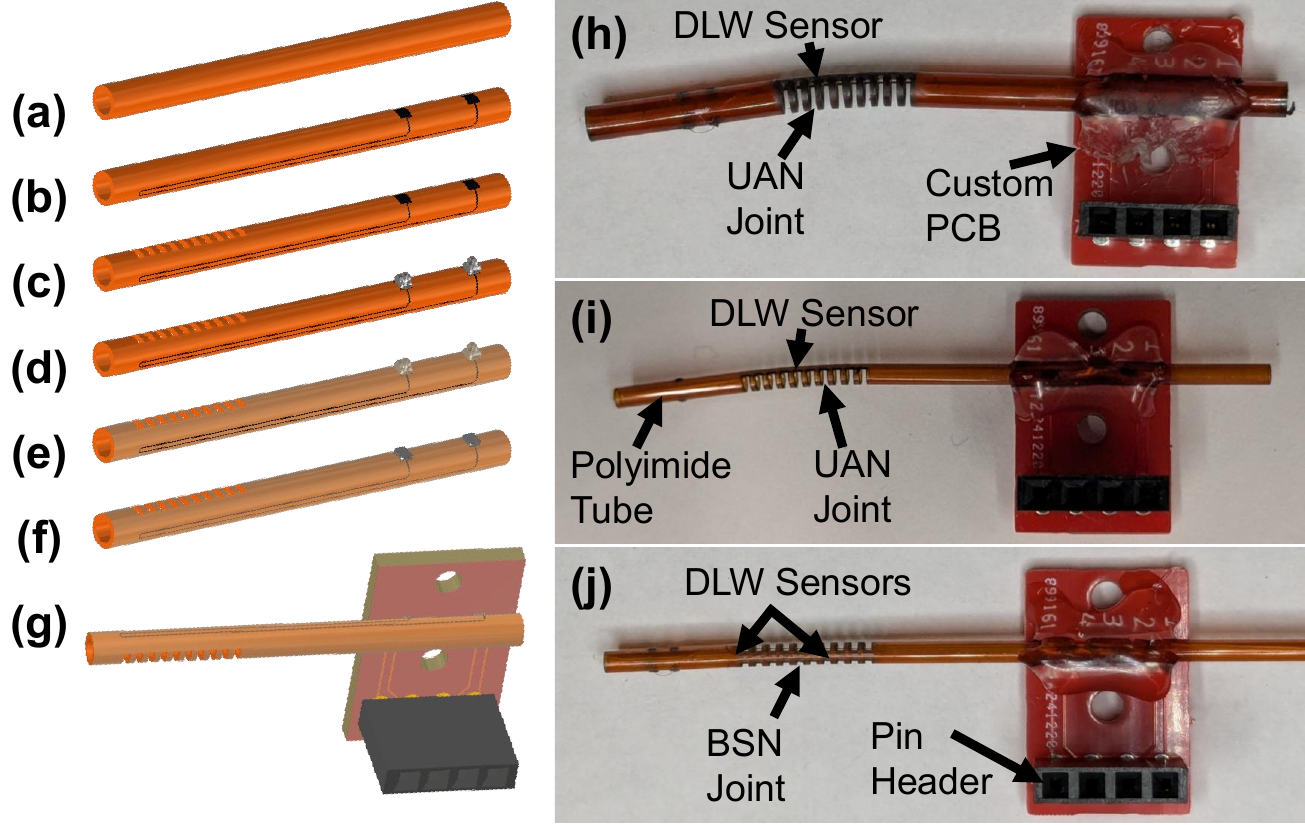}
\caption{The sensing joint fabrication process (a) begins with stock polyimide tubing. (b) A laser is used to draw the sensor on the tube and (c) cut notches in the tube. (d) The contact pads are protected by silver epoxy before (e) the entire structure is coated with parylene-C. (f) The parylene is sanded off the contact pads, and (g) additional silver epoxy is used to assemble the sensing joint with the PCB. The images of the completed sensing joints: (h) Joint A, (i) Joint B, and (j) Joint C.}
\label{fig_assembly}
\end{figure}

\subsubsection{Laser Micromachining}

\begin{table}[tb]
\begin{center}
\caption{Design Parameters}\label{table_tube_params}
\begin{tabular}{|c || c | c | c |} 
\hline
& \textbf{Joint A} & \textbf{Joint B} & \textbf{Joint C} \\ 
\hline\hline
Notch Style & UAN & UAN & BSN \\ 
\hline
Outer Radius, $r_o$ & \SI{1.24}{\milli\meter} & \SI{0.800}{\milli\meter} & \SI{0.800}{\milli\meter} \\ 
\hline
Inner Radius, $r_i$ & \SI{0.875}{\milli\meter} & \SI{0.610}{\milli\meter} & \SI{0.610}{\milli\meter}\\
\hline
Notch Circumference, $s_n$ & \SI{5.61}{\milli\meter} & \SI{3.52}{\milli\meter} & \SI{1.65}{\milli\meter} $\times 2$\\
\hline
Notch Width, $h$ & \SI{0.500}{\milli\meter} & \SI{0.500}{\milli\meter} & \SI{0.500}{\milli\meter}\\
\hline
Notch Spacing, $c$ & \SI{0.500}{\milli\meter} & \SI{0.500}{\milli\meter} & \SI{0.500}{\milli\meter}\\
\hline
Number of Notches, $N$ & \SI{10}{} & \SI{10}{} & \SI{10}{} $\times 2$\\
\hline
Joint Length, $L$ & \SI{9.5}{\milli\meter} & \SI{9.5}{\milli\meter} & \SI{9.5}{\milli\meter}\\
\hline
Sensor Circumference, $s_s$ & \SI{0.400}{\milli\meter} & \SI{0.400}{\milli\meter} & \SI{2.13}{\milli\meter} $\times 2$ \\
\hline
\end{tabular}
\end{center}
\end{table}

A piece of machine steel of diameter \SI{1.2}{\milli\meter} or \SI{1.5}{\milli\meter} was placed inside the \SI{1.6}{\milli\meter} or \SI{2.5}{\milli\meter} outer diameter stock polyimide tubing (Fig.\,\ref{fig_assembly}(a)), respectively, to increase the stiffness of the polymer tubes during the machining process and secured on one end using a small amount of cyanoacrylate glue. The stock was then fixtured in the femtosecond laser micromachining system (WS-Flex Ultra-Short Pulse Laser Workstation, Optec, Frameries, Belgium), with the right end secured in the collet of the system's rotation stage and the left end resting in a custom 3D-printed jig. The overall parameters of the laser system used are given in Table\,\ref{table_laser_params}. Parameters were selected for fabricating the sensor and notches based on previous work \cite{Rothe2025TowardsSensors, Malhotra2025DesignApplications}, as well as calibration of the laser. First, the sensing elements were machined using the settings given in Table\,\ref{table_laser_params}, column (a) (Fig.\,\ref{fig_assembly}(b)). Then, without changing the laser setup, the notches were machined using the settings given in Table\,\ref{table_laser_params}, column (b) (Fig.\,\ref{fig_assembly}(c)). Finally, a clean cut was made to separate the sensing joint from the remainder of the tube using the same settings. Fabricating both the sensor and the notches using the same equipment and setup eliminates the possibility of human error in aligning or assembling the sensor with the notches. 

\begin{table}[t!]
\begin{center}
\caption{Laser Parameters}\label{table_laser_params}
\begin{tabular}{|c || c | c |} 
\hline
& \textbf{(a) Sensor} & \textbf{(b) Notches} \\ 
\hline
\hline
Wavelength, $\lambda$ & \SI{1030}{\nano\meter} & \SI{1030}{\nano\meter} \\ 
\hline
Power, $P$ & \SI{160}{\milli\watt} & \SI{2.00}{\watt} \\ 
\hline
Scan Speed, $v$ & \SI{1.00}{\milli\meter\per\second} & \SI{25}{\milli\meter\per\second} \\ 
\hline
Pulse Frequency, $f$ & \SI{200}{\kilo\hertz} & \SI{60}{\kilo\hertz} \\ 
\hline
Pulse Width, $t_p$ & \SI{550}{\femto\second} & \SI{215}{\femto\second} \\ 
\hline
Repetitions, $n_{\mathrm{rep}}$ & \SI{1}{} & \SI{32}{} \\ 
\hline
\end{tabular}
\end{center}
\end{table}

\subsubsection{Protection of the Contact Pads}
To protect the contact pads during the subsequent steps of the process, a small amount of conductive silver epoxy (8331D, MG Chemicals, Burlington, ON, Canada) was placed on each contact pad and allowed to cure overnight (Fig.\,\ref{fig_assembly}(d)). 

\subsubsection{Parylene Coating}
To seal the sensor and protect it during use, the joints were coated with parylene-C (Galentis, Venice, Italy). A Labcoater PDS 2010 (Specialty Coating Systems, Indianapolis, IN, USA) was used to perform the vacuum deposition of \SI{8}{\gram} of parylene, for a nominal expected thickness of approximately \SI{5}{\micro\meter} (Fig.\,\ref{fig_assembly}(e)). 

\subsubsection{PCB Assembly}
The parylene coating was carefully sanded off of the protected contact pads using a small file (Fig.\,\ref{fig_assembly}(f)). Then, additional silver epoxy was used to attach the contact pads to a custom printed circuit board (PCB), which provided female pin connections for wiring the sensor to the other system electronics (Fig.\,\ref{fig_assembly}(g)). A custom 3D-printed jig was used to align the contact pads on the sensing joint with the contact pads on the PCB. The epoxy was allowed to dry overnight. Finally, hot melt adhesive was used to improve the mechanical connection between the joint and the PCB, as the epoxy provides a strong electrical connection but a weak mechanical connection. The three designed joints, Joint A, Joint B, and Joint C, are shown in Fig.\,\ref{fig_assembly}(h), Fig.\,\ref{fig_assembly}(i), and Fig.\,\ref{fig_assembly}(j), respectively.

\subsubsection{System Assembly}
A nylon tendon of diameter \SI{0.28}{\milli\meter} (OmniFlex 8lb, Zebco, Tulsa, OK, USA) was affixed to the distal end of the sensing joint. The sensing joint-PCB assembly was clamped firmly with screws to the actuation system, while the tendon was attached to a lead screw actuated by a DC motor. The sensors were attached to a simple voltage divider circuit, with the output connected to a data acquisition board (National Instruments, Austin, TX, USA). The constant resistor in the voltage divider circuit had a measured resistance of \SI{99.6}{\kilo\ohm}. For Joint C, which had two sensors, the value of the second resistor was \SI{102.5}{\kilo\ohm}. An electromagnetic (EM) tracker (Northern Digital Inc. Aurora, Waterloo, ON, Canada) was placed in the tip of the sensing joint to provide ground-truth deflection angle data. All data during the experiments was collected at \SI{500}{\hertz}.

\section{Calculations and Modeling}\label{section_modeling}
\subsection{Strain}\label{section_strain}

\begin{figure}[tb]
\centering
\includegraphics[width=0.99\columnwidth]{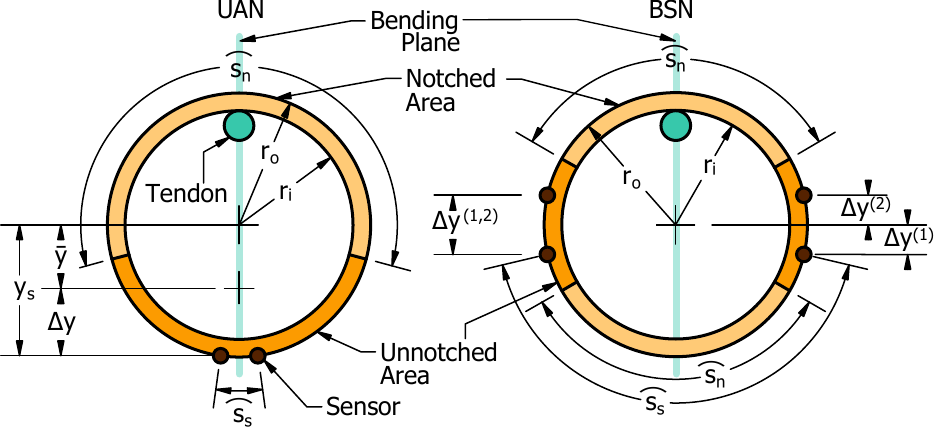}
\caption{Schematic of the notch cross section for UAN and BSN designs. }
\label{fig_tube_cross_section}
\end{figure}

The strain, $\varepsilon$, experienced in the sensor due to a joint deflection of angle $\theta$ can be calculated using the geometry of the joint. It is assumed that bending in the unnotched sections is negligible \cite{Rothe2024Model-basedBAL}. The perpendicular distance between the sensors and the center of the tube, $y_s$, is given by:
\begin{equation}
    y_s = r_o \mathrm{cos}\left({\phi_s}/{2}\right),\quad\mathrm{where}\quad \phi_s = {s_s}/{r_o}
\end{equation}
where $s_s$ is the circumferential distance between the arms of the sensor and $r_o$ is the outer radius of the tube. The distance between the neutral plane and the center of the tube, $\overline{y}$ (Fig.\,\ref{fig_tube_cross_section}), depends on the UAN or BSN notch style \cite{Rothe2025TowardsSensors}:
\begin{equation}
\overline{y}_A = \overline{y}_B = \frac{4\sin\left(\pi - {s_n}/{(2r_o)}\right)\left(r_o^3 - r_i^3\right)}
{3\left(2\pi - {s_n}/{r_o}\right)\left(r_o^2 - r_i^2\right)}, \quad
\overline{y}_C = 0
\end{equation}
where $r_i$ is the inner radius of the tube and $s_n$ is the arc length of the notch. The strain in the sensor, $\varepsilon$, is given by:
\begin{equation} \label{eq_strain}
    \varepsilon = \left(\Delta y \theta\right)/{L}, \quad \quad \Delta y = y_s - \overline{y}
\end{equation}
where $\Delta y$ is the distance between the neutral plane and the plane of the sensor.
\subsection{Sensor Resistance}\label{section_resistance}
A voltage divider circuit is used to obtain the sensor readings. When the sensed voltage is $V$, the input voltage is $V^+$, and the known resistor has a resistance of $R_k$, the resistance in the sensor, $R$ is given by:
\begin{equation}\label{eq_resistance}
    R = R_k/\left({\left(V^+ / V\right) - 1}\right)
\end{equation}

\subsection{Linear Model}\label{section_linear}
A simple linear model has been used to characterize DLW strain gauges in the literature \cite{Barja2024Laser-InducedMonitoring, Carvalho2018LaserInducedPolyimide}. In such models, the gauge factor, \(\mathrm{GF}\), is used to express how resistance, \(R\), changes in response to strain, \(\varepsilon\), and is defined by \cite{Beckwith1982MechanicalMeasurements}:
\begin{equation} \label{eq_gf}
    \mathrm{GF} = (R - R_0)/(R_0\varepsilon)
\end{equation}
where \(R_0\) is the initial resistance of the unstrained gauge and \(R\) is the resistance associated with strain $\varepsilon$. Re-arranging this equation yields the forward and inverse relationship between strain and sensor resistance, given by:
\begin{equation}\label{eq_onesensor_angle}
    R = R_0 \mathrm{GF} \varepsilon + R_0, \quad \quad \varepsilon = (R - R_0)/\left(R_0 \mathrm{GF}\right)
\end{equation}
In a joint with two strain gauges of gauge factors \(\mathrm{GF}^{(1)}\) and \(\mathrm{GF}^{(2)}\) and initial resistances of \(R^{(1)}_0\) and \(R^{(2)}_0\), the deflection angle can be estimated independently from each gauge using the above equations. Alternatively, the deflection angle may be calculated from both gauges, where \(R^{(1)}\) and \(R^{(2)}\) are the resistance of each gauge and $\Delta y^{(1,2)}$ is the distance between the planes of the gauges (Fig.\,\ref{fig_tube_cross_section}):
\begin{equation}\label{eq_twosensor_angle}
    \theta = \frac{L}{\Delta y^{(1,2)}}\left(\frac{R^{(2)} - R^{(2)}_0}{R^{(2)}_0\mathrm{GF}^{(2)}} - \frac{R^{(1)} - R^{(1)}_0}{R^{(1)}_0\mathrm{GF}^{(1)}}\right)
\end{equation}

\subsection{Prandtl-Ishlinskii Model}\label{section_pi}
Other works have noted that the relationship between strain and resistance for DLW sensors may not be linear, and have modeled it using various piecewise linear \cite{Wang2025ADetection, Huang2020WearableRubber}, 
and quadratic \cite{Chen2025PiezoresistiveStructures} models. In this work, we propose using the generalized Prandtl-Ishlinskii (PI) hysteresis model \cite{AlJanaideh2023TheNotes} to represent the relationship between strain and sensor resistance. The PI model is capable of capturing both nonlinearity and hysteresis. 
The forward model estimates the sensor resistance, $R(t)$, given the sensor strain, $\varepsilon(t)$ through time, $t$. The PI model used in this work is based on the play operator, $\Gamma_{\rho}(\nu(t))$, given by \cite{AlJanaideh2023TheNotes}:

\small
\begin{equation}
    \Gamma_{\rho}(\nu(t))= \mathrm{max}\left(\nu(t)-\rho, \mathrm{min}\left(\nu(t) + \rho, \Gamma_{\rho}(\nu(t-1))\right)\right)
\end{equation}
\normalsize

\noindent with the initial condition given by:
\begin{equation}
    \Gamma_{\rho}(\nu(0)) = \mathrm{max}\left(\nu(0)-\rho, \mathrm{min}\left(\nu(0) + \rho, M_0\right)\right)
\end{equation}
Where $\rho$ is the radius of the play operator, $t$ is time, $M_0$ is the initial value, and $\nu(t)$ is the input to the play operator. 

The PI model allows the play operator to be combined with a function that modifies its shape. If the data appears to saturate, a hyperbolic tangent function is a good choice for this. Thus, the input to the play operator for the forward model is the hyperbolic tangent of the strain:

\vspace{-0.25cm}
\small
\begin{equation}
    \nu(\varepsilon(t)) = \begin{cases} a_1 \mathrm{tanh}(a_2\varepsilon(t) + a_3) + a_4, & \varepsilon(t) \geq \varepsilon(t-1)\\ a_5 \mathrm{tanh}(a_6\varepsilon(t) + a_7) + a_8, & \varepsilon(t) < \varepsilon(t-1)
\end{cases}
\end{equation}
\normalsize

\noindent where $a_j$, $j=\{1,2,\dots,8\}$ are constants. The expected sensor resistance is computed as a weighted sum of $N$ play operators, with weights $P_i$ and radii $\rho_i$, $i=\{1,2,\dots,N\}$, where $R_c$ and $\varepsilon_c$ are constants, given by:
\begin{equation}
    R(t) = \sum\nolimits_{i=1}^{N} P_i \Gamma_{\rho_i} (\nu(\varepsilon(t) - \varepsilon_c)) + R_c
\end{equation}

The inverse model estimates the sensor strain, $\varepsilon(t)$, given the measured sensor resistance, $R(t)$. The inverse PI model incorporates another PI model of order $N$  \cite{AlJanaideh2023TheNotes,Bhore2018IdentificationNonlinearities}: %
\begin{equation}
    \varepsilon(t) = \nu^{-1}\left(\sum\nolimits_{i=1}^{N} Q_i \Gamma_{\zeta_i} (R(t) - R_c) \right)  + \varepsilon_c
\end{equation}
where the the radii, $\zeta_i$, $i=\{1,2,\dots,N\}$, and the weights $Q_i$, $i=\{1,2,\dots,N\}$, are given by:

\vspace{-0.25cm}
\small
\begin{equation}
    \begin{split}
        \zeta_i = & \sum\nolimits_{j=1}^{i} P_j(\rho_i- \rho_j), \quad\quad\quad Q_1 = {1}/{P_1} \\
        Q_i = & {-P_i}/\left({\left(P_1 + \sum\nolimits_{j=1}^{i} P_j \right)\left(P_1 + \sum\nolimits_{j=1}^{i-1} P_j \right)}\right) 
    \end{split}
\end{equation}
\normalsize

\noindent and $\nu^{-1}(\mu(t))$ is the inverse of the hyperbolic tangent function for input $\mu(t)$, given by:

\vspace{-0.25cm}
\small
\begin{equation}
    \nu^{-1}(\mu(t)) = \begin{cases}
    &\frac{1}{a_2}\left(\mathrm{atanh}\left(\frac{1}{a_1}(\mu(t)-a_4)\right)-a_3\right), \\ & \hfill \mu(t) \geq \mu(t-1) \\
    &\frac{1}{a_6}\left(\mathrm{atanh}\left(\frac{1}{a_5}(\mu(t)-a_8)\right)-a_7\right), \\ & \hfill \mu(t) < \mu(t-1)
\end{cases}
\end{equation}
\normalsize

\section{Experiments and Results}\label{section_experiments}
Several experiments were conducted to characterize the behavior and performance of the sensing joints, such as the hysteresis of the sensor, rate dependence of the sensor, and relaxation of the joint. The models described in Section\,\ref{section_modeling} were fit to the experimental data and used to perform closed-loop control during the joint's motion. Finally, a demonstration of a sensing joint in a hydrogel phantom was performed to show the feasibility of the proposed approach in a physiological environment. 

\subsection{Sensor Characterization}\label{section_characterization}
\subsubsection{Resistance-Strain Relationship}\label{section_rich}

\begin{figure}[tb]
\centering
\includegraphics[width=0.99\columnwidth]{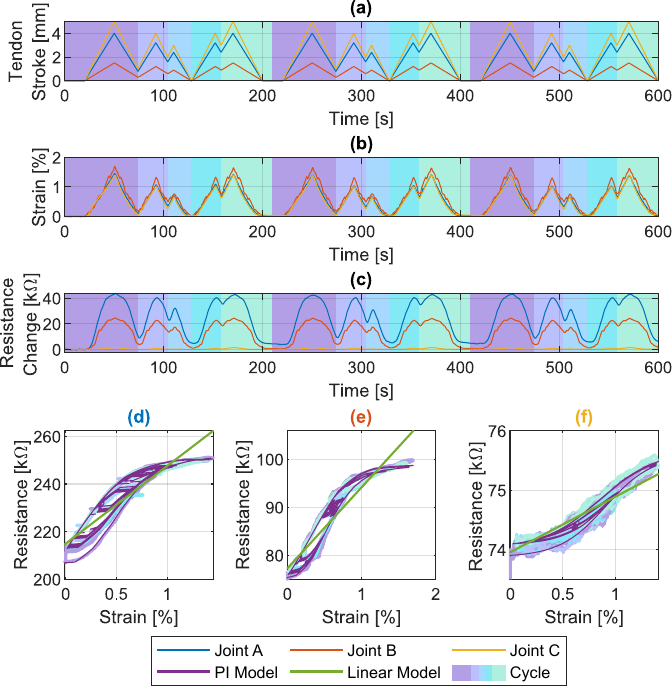}
\caption{Data for the sensor characterization experiment. (a) The tendon stroke pattern. This pattern was repeated five times in the experiment. (b) The strain, calculated from the ground truth angle. (c) The change in sensor resistance, calculated from the sensor voltage. The relationship between sensor resistance and strain for (d) Joint A, (e) Joint B, and (f) Joint C. Each distinct loading/unloading cycle is shaded in a different color.}
\label{fig_rich}
\vspace{-0.5cm}
\end{figure}

To characterize the relationship between resistance change and strain for the three joints, the tendon was pulled and released with the profile shown in Fig.\,\ref{fig_rich}(a), consisting of five cycles of different amplitudes. The ground truth sensor deflection angle was recorded using the electromagnetic tracker, and the strain at the sensor was calculated using Eq.\,\eqref{eq_strain}, shown in Fig.\,\ref{fig_rich}(b). Overall, the range of strains tested was between \SI{0}{\percent} and approximately \SI{1.5}{\percent}, which corresponded to about \SI{26}{\degree}, \SI{58}{\degree}, and \SI{40} {\degree} bending in Joints A, B, and C, respectively. For Joint C, which has two sensors, only the sensor in tension is considered in this section. The other sensor is considered in Section\,\ref{section_compression}. The sensor voltage was also recorded, and its resistance calculated using Eq.\,\eqref{eq_resistance} is shown in Fig.\,\ref{fig_rich}(c). For this and the other experiments, all input signals were filtered with moving average filters of 500 samples. This is relatively aggressive filtering intended to counter the high noise observed in the system signals. In future work, other noise mitigation strategies may be attempted. A simple linear best-fit model for the resistance-strain relationship was found using regression. The gauge factor and initial resistance (intercept) are shown in Table\,\ref{table_model_params}. 

 We observed that the gauge factor for Joint C is lower than the others, indicating that the sensor on Joint C is less sensitive. Joint A and Joint B have unidirectional notches, allowing the sensor to be placed very close to the plane of bending (Fig.\,\ref{fig_tube_cross_section}). On the other hand, Joint C has symmetrical notches, thus the sensors must be placed nearly \SI{90}{\degree} around the tube from the bending plane (Fig.\,\ref{fig_tube_cross_section}). We hypothesize that this difference causes the large variation in gauge factor between the two designs. The change in resistivity of graphene is related to microscopic cracking of the conductive network \cite{Thakur2025EngineeredSensors}. In the BSN design, the graphene is constrained by the adjacent raw polyimide as it bends nearly tangentially to the surface of the tube. However, the sensor in the UAN design bends nearly orthogonally to the tube's surface; therefore, the graphene network is unconstrained, allowing larger cracks to form in the graphene network. Future research will be necessary to fully understand the phenomenon.  

Although the linear model may be adequate for some applications, the relationship between resistance and strain is both hysteretic and nonlinear. Therefore, the PI model discussed in Section\,\ref{section_pi} was fit to the data. The order was chosen to be $N = 10$. The $N$ radii, $\rho_i$ were chosen to be evenly spaced from 0 to $\rho_{\mathrm{max}}$. Because the PI model discussed in Section\,\ref{section_pi} is centered around the origin, it was first necessary to shift the data to center it on the origin. The resistance at the center, $R_c$, was estimated as the resistance where the slope of the resistance-strain plot was greatest during the loading phase of the first cycle, with a value of $m_c$. The strain at the center, $\varepsilon_c$, was estimated as the mean of the corresponding strain in the loading phase of the first cycle and the unloading phase of the last cycle. The weights, $P_i$, and constants, $a_j$, were chosen using nonlinear optimization. Because the optimization was highly sensitive to its initial conditions, a good starting ``guess'' of these values was necessary. The initial weights, $P^{(0)}_i$ and initial constants, $a^{(0)}_j$, as well as $\rho_{\mathrm{max}}$, are given by:
\begin{equation}
    \begin{split}
        a^{(0)}_1 &= a^{(0)}_5 = R_{\mathrm{peak}} - R_c \\
        a^{(0)}_2 &= a^{(0)}_6 = m_c/\left(2(R_{\mathrm{peak}} - R_c)\right) \\
        a^{(0)}_3 &= a^{(0)}_4 = a^{(0)}_7 = a^{(0)}_8 = 0 \\
        P^{(0)}_i & = \left(R_{\mathrm{peak}} - R_c\right)/{N}, i = {1, 2, \dots, N} \\
        \rho_{\mathrm{max}} & = a^{(0)}_1 \mathrm{tanh}\left(a^{(0)}_2\varepsilon_{\mathrm{peak}} + a^{(0)}_3\right) + a^{(0)}_4
    \end{split}
\end{equation}
The PI model is shown in a purple color while the linear model is shown in a green color in Fig.\,\ref{fig_rich}(d), Fig.\,\ref{fig_rich}(e), and Fig.\,\ref{fig_rich}(f) for Joint A, Joint B, and Joint C, respectively. Selected model parameters are given in Table\,\ref{table_model_params}.

\subsubsection{Rate Dependence of Sensor Response}\label{section_multispeed}

\begin{table}[tb]
\begin{center}
\caption{Angle Estimation Parameters}\label{table_model_params}
\begin{tabular}{| c || c || c | c | c|} 
\hline
\multicolumn{2}{|c|}{}& \textbf{Joint A} & \textbf{Joint B} & \textbf{Joint C} \\ 
\hline\hline
\multirow{2}{*}{Linear Model} & $\mathrm{GF}$ & \SI{15.3}{} & \SI{22.0}{} & \SI{1.28}{} \\ 
\cline{2-5}
& $R_0$ & \SI{215}{\kilo\ohm} & \SI{77.2}{\kilo\ohm} & \SI{74.0}{\kilo\ohm} \\ 
\hline
\hline
\multirow{5}{*}{PI Model} & $R_c$ & \SI{224}{\kilo\ohm} & \SI{87.9}{\kilo\ohm} & \SI{74.7}{\kilo\ohm} \\ 
\cline{2-5}
& $\varepsilon_c$ & \SI{0.307}{\percent} & \SI{0.531}{\percent} & \SI{0.804}{\percent} \\ 
\cline{2-5}
& $m_c$ & \SI{5136}{\kilo\ohm} & \SI{3090}{\kilo\ohm} & \SI{181}{\kilo\ohm} \\ 
\cline{2-5}
& $R_{\textrm{peak}}$ & \SI{250.}{\kilo\ohm} & \SI{99.2}{\kilo\ohm} & \SI{754}{\kilo\ohm} \\ 
\cline{2-5}
& $\varepsilon_{\textrm{peak}}$ & \SI{1.35}{\percent} & \SI{1.57}{\percent} & \SI{1.37}{\percent} \\ 
\hline
\end{tabular}
\vspace{-0.5cm}
\end{center}
\end{table}

\begin{figure}[tb]
\centering
\includegraphics[width=0.99\columnwidth]{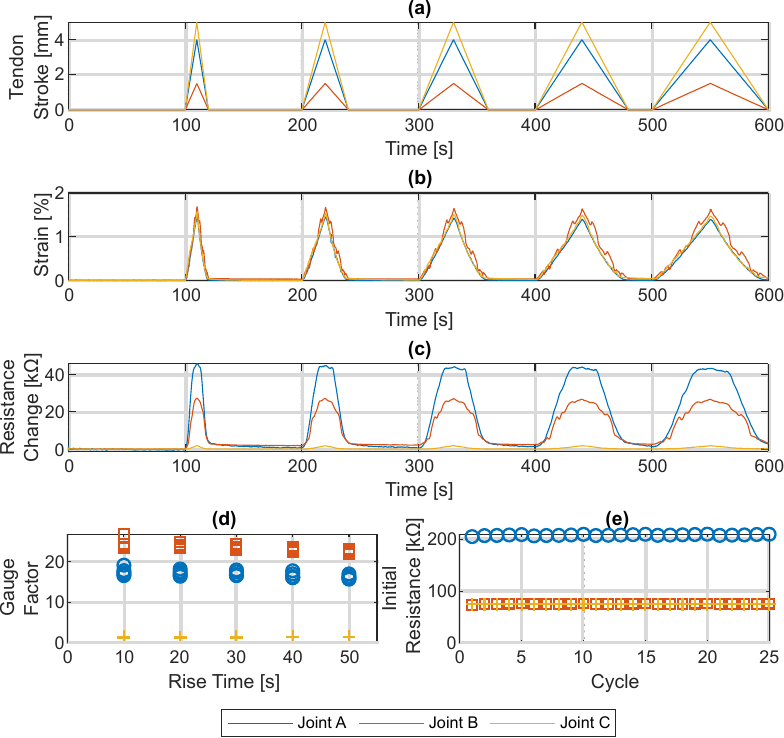}
\caption{Data for the rate-dependence experiment. (a) The tendon stroke pattern, featuring five cycles of loading and unloading with identical amplitude but different rise and fall times. (b) The strain, calculated from the ground truth measured angle. (c) The change in sensor resistance, calculated from the measured sensor voltage. (d) The linear gauge factor, $\mathrm{GF}$, of each cycle, plotted against the rise time for that cycle (e) The initial resistance, $R_0$, of each cycle, plotted against the cycle number. }
\label{fig_multispeed}
\vspace{-0.25cm}
\end{figure}

Some polymeric materials exhibit rate-dependent behavior \cite{Krempl2003RatePolymers}; therefore, it was important to determine if the sensor's response is affected by the speed of the joint's actuation. To this end, the tendon of each joint was pulled and released with the profile shown in Fig.\,\ref{fig_multispeed}(a), consisting of five cycles. The amplitude of the tendon stroke was the same for each cycle; however, the duration of the tendon stroke was varied between \SI{10}{\second} and \SI{50}{\second} to achieve a variation in stroke speed. The cycles were repeated five times over a total time of \SI{3000}{s}. The sensor strain and resistance were obtained as in Section\,\ref{section_rich} and shown in Fig.\,\ref{fig_multispeed}(b) and Fig.\,\ref{fig_multispeed}(c), respectively. For each cycle, the linear gauge factor was determined by fitting the linear model to each individual cycle, and plotted in Fig.\,\ref{fig_multispeed}(d). The correlation coefficient between stroke duration and gauge factor was calculated to be \SI{0.0062}{}, \SI{0.0010}{}, and \SI{0.0011}{} for Joint A, Joint B, and Joint C, respectively, indicating low correlation between speed and sensor response. 

In addition, the initial resistance of each cycle, $R_0$, which is the intercept of the linear model, is plotted Fig.\,\ref{fig_multispeed}(e). The standard deviation of the initial resistance values was calculated to be \SI{1.10}{\kilo\ohm}, \SI{0.596}{\kilo\ohm}, and \SI{0.038}{\kilo\ohm} for Joint A, Joint B, and Joint C, respectively, indicating that the drift in the sensor across the 25 total cycles was low. 

\subsubsection{Relaxation}\label{section_creep}

\begin{figure}[tb]
\centering
\includegraphics[width=0.99\columnwidth]{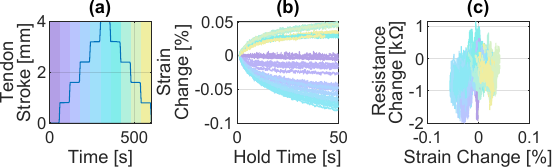}
\caption{Data for the relaxation experiment for Joint A. (a) The tendon stroke pattern, featuring stepwise loading and unloading. This pattern was repeated five times in the experiment (b) The decay in strain during each period of constant tendon stroke. (c) The relationship between resistance and strain during each period of constant tendon stroke.}
\label{fig_creep}
\end{figure}
In another experiment, the tendon stroke was held constant at various amplitudes, shown in Fig.\,\ref{fig_creep}(a) for Joint A. This stroke pattern was repeated five times over \SI{3000}{\second} while the strain and sensor resistance were recorded. It was observed that when the tendon stroke was held constant for a time, the joint angle and therefore calculated strain changed, presumably due to viscoelastic properties in the polymer tube and polymer tendon (Fig.\,\ref{fig_creep}(b)). However, during this time, the change in sensor resistance (Fig.\,\ref{fig_creep}(c)) was small and not strongly correlated to the change in strain, with a correlation coefficient of \SI{0.118}{}. Thus, the proposed sensor may be inadequate for capturing viscoelastic behavior and may need to be used in conjunction with other sensing methods when these effects are expected to be significant.  

\subsection{Angle Estimation}\label{section_reconstruction} 

\begin{figure}[tb]
\centering
\includegraphics[width=0.99\columnwidth]{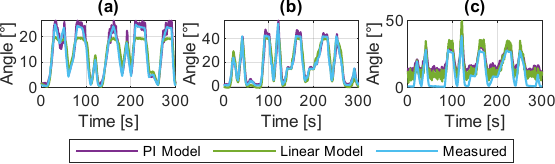}
\caption{The ground truth angle and the angle estimated using the linear and PI models for (a) Joint A, (b) Joint B, and (c) Joint C.}
\label{fig_reconstruction}
\end{figure}
\subsubsection{Angle Estimation with One Sensor}\label{section_estimation_onesensor}

\begin{table}[tb]
\begin{center}
\caption{Angle Estimation Results}\label{table_reconstruction_error}
\begin{tabular}{| c || c | c | c|} 
\hline
& \textbf{Joint A} & \textbf{Joint B} & \textbf{Joint C} \\ 
\hline\hline
Linear & \SI{2.76}{\degree} & \SI{3.51}{\degree} & \SI{5.25}{\degree} \\ 
\hline
PI & \SI{1.76}{\degree} & \SI{2.65}{\degree} & \SI{7.57}{\degree} \\
\hline
\end{tabular}
\vspace{-0.5cm}
\end{center}
\end{table}

The inverse linear and PI models for the three joints were calculated based on the forward models determined in Section\,\ref{section_rich} according to Section\,\ref{section_linear} and Section\,\ref{section_pi}, respectively. The tendon was actuated in a randomly generated pattern, which was repeated for a total of \SI{3000}{\second}. The recorded sensor resistance was used to predict the strain using each of the inverse models. Because the $\mathrm{atanh}$ function has asymptotic behavior, extrapolating beyond the range of the original data collected is unlikely to be accurate. Therefore, all resistances above and below this range were replaced with the maximum and minimum resistance observed during the sensor characterization experiment performed in Section\,\ref{section_rich} for the PI model prediction. Then, the expected joint angle was calculated using Eq.\,\eqref{eq_strain}. Because the $\mathrm{atanh}$ function is very steep near its asymptotes, the model is highly sensitive to small changes in resistance. Therefore, the predicted angles were filtered aggressively with a moving average filter of 1000 samples. The ground truth angle, linear model predicted angle, and PI model predicted angle for Joint A, Joint B, and Joint C are shown in Fig.\,\ref{fig_reconstruction}(a), Fig.\,\ref{fig_reconstruction}(b), and Fig.\,\ref{fig_reconstruction}(c), respectively. The root mean square errors (RMSE) between the predicted and actual angles are recorded in Table\,\ref{table_reconstruction_error}. The PI model was more accurate for Joint A and B, while the linear model remained more accurate for Joint C. As previously discussed, Joint C is also the least sensitive potentially due to the difference in the direction of the graphene cracking; it is observed that in this configuration, the hysteretic and saturating behavior captured by the PI model is less significant. 

\subsubsection{Angle Estimation with Two Sensors}\label{section_compression}

\begin{figure}[tb]
\centering
\includegraphics[width=0.95\columnwidth]{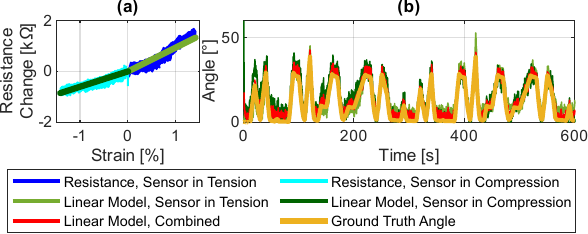}
\caption{(a) The modeled and actual resistance versus strain for the sensor in tension and compression. (b) The ground truth and estimated angle based on the sensor in tension, the sensor in compression, and both. }
\label{fig_twosensor}
\end{figure}

\begin{table}[tb]
\begin{center}
\caption{Two Sensor Model Parameters}\label{table_twosensor_coefficients}
\begin{tabular}{| c || c | c | c |} 
\hline
& \textbf{Gauge} & \textbf{Initial} & \textbf{Correlation} \\ 
& \textbf{Factor} & \textbf{Resistance} & \textbf{Coefficient} \\ 
& \textbf{$(\mathrm{GF})$} & \textbf{$(R_0)$} & \textbf{$(r^2)$}\\ 
\hline\hline
Sensor in Tension & \SI{1.28}{} & \SI{74.0}{\kilo\ohm} & \SI{0.888}{} \\ 
\hline
Sensor in Compression & \SI{0.680}{} & \SI{88.9}{\kilo\ohm} & \SI{0.892}{} \\ 
\hline
\end{tabular}
\vspace{-0.25cm}
\end{center}
\end{table}

In the case where there are two sensors on the same joint, as in Joint C, which has the BSN design shown in Fig.\,\ref{fig_tube_cross_section}, it is possible to combine the readings from both sensors to obtain a more accurate and robust estimation of the joint's angle. During the experiment described in Section\,\ref{section_rich}, the resistance of the sensor in compression was also recorded. Because the linear model proved more accurate for Joint C in Section\,\ref{section_reconstruction}, this model was chosen for the two sensor angle estimation. A separate linear model was fit for the sensor in tension and the sensor in compression (Fig.\,\ref{fig_twosensor}(a)) based on Eq.\,\eqref{eq_onesensor_angle}. The values are given in Table\,\ref{table_twosensor_coefficients}, along with the correlation coefficients. The fitted models were used individually predict the joint angle as in Section\,\ref{section_reconstruction}. As stated in that section, the sensor in tension achieved an RMSE of \SI{5.25}{\degree}. The sensor in compression achieved an RMSE of \SI{3.84}{\degree}. The characteristics and resistance of both sensors together were then used to predict the joint angle as well using Eq.\,\eqref{eq_twosensor_angle}, achieving an RMSE of \SI{3.13}{\degree}. The results for a portion of the \SI{3000}{\second} total experiment are shown in Fig.\,\ref{fig_twosensor}(b). 

\subsection{Joint Control}\label{section_control}

\begin{figure}[t!]
\centering
\includegraphics[width=0.8\columnwidth]{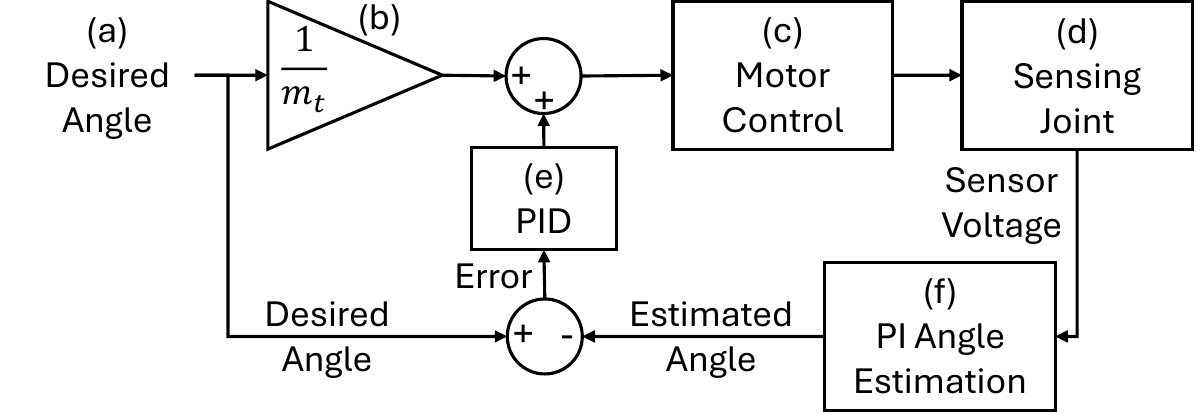}
\caption{Closed-loop control system for Joint D. (a) Desired angle input. (b) Tendon stroke-joint angle model. (c) Low level motor control. (d) Physical sensing joint. (e) PID controller. (f) Angle estimation subsystem. }
\label{fig_control_system}
\end{figure}

\begin{figure}[t]
\centering
\includegraphics[width=0.95\columnwidth]{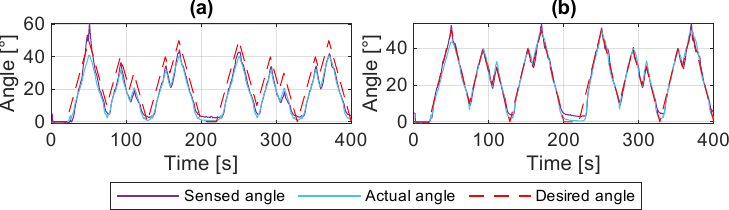}
\caption{Desired angle, actual angle, and estimated angle based on sensor input for (a) open-loop control and (b) closed-loop control.}
\label{fig_control}
\end{figure}

\begin{table}[t]
\begin{center}
\caption{Joint Control Results}\label{table_control_results}
\begin{tabular}{| c || c | c |} 
\hline
& \textbf{Open-Loop} & \textbf{Closed-Loop} \\ 
& \textbf{Control} & \textbf{Control} \\ 
\hline\hline
Sensed vs Actual Angle RMSE & \SI{1.88}{\degree} & \SI{2.29}{\degree} \\ 
\hline
Actual vs Desired Angle RMSE& \SI{8.14}{\degree} & \SI{2.92}{\degree} \\ 
\hline
Sensed vs Desired Angle RMSE& \SI{8.46}{\degree} & \SI{2.93}{\degree} \\ 
\hline
\end{tabular}
\end{center}
\end{table}

\begin{table}[t]
\begin{center}
\caption{Demonstration Results}\label{table_demo_results}
\begin{tabular}{| c || c | c |} 
\hline
& \textbf{Linear Model} & \textbf{PI Model} \\
\hline\hline
Trial 1 RMSE& \SI{7.11}{\degree} & \SI{8.17}{\degree} \\ 
\hline
Trial 2 RMSE& \SI{4.51}{\degree} & \SI{4.51}{\degree} \\ 
\hline
Trial 3 RMSE& \SI{4.25}{\degree} & \SI{4.18}{\degree} \\ 
\hline
\end{tabular}
\end{center}
\end{table}

\begin{figure}[t]
\centering
\includegraphics[width=0.90\columnwidth]{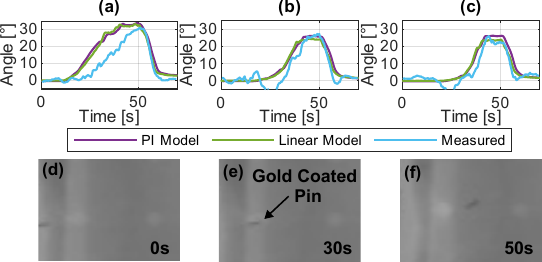}
\caption{Data recorded during the hydrogel phantom demonstration for (a) Trial 1, (b) Trial 2, and (c) Trial 3. Sample fluoroscopy images recorded during Trial 2 at timestamps of (d) \SI{0}{\second}, (e) \SI{30}{\second}, and (f) \SI{50}{\second}}
\label{fig_phantom_demo}
\end{figure}

Without including sensing feedback, open-loop control of a robotic joint is possible. To assess the performance of open-loop control of the robotic joints, an experiment was performed using Joint D. Joint D was a unidirectional asymmetric notch joint with design identical to Joint B. The PI model for Joint D was determined using the procedure in Section\,\ref{section_rich}, with ($N$=\SI{3}{}) for the purpose of simplifying the control system. A simple, linear, empirical model relating tendon stroke to joint angle was determined by finding the best fit slope relating these quantities, found to be $m_t$=\SI{35.49}{\degree\per\milli\meter}. An open-loop control scheme was implemented in Simulink (MathWorks, Natick, MA, USA), where $1/m_t$ was the gain between the desired angle and the commanded tendon stroke given as an input into the low level motor control. The actual ground-truth angle recorded by the EM tracking system and the sensed angle (unused by the controller) was computed according to Section\,\ref{section_reconstruction}. These are shown in Fig.\,\ref{fig_control}(a) and the RMSE between them are given in Table\,\ref{table_control_results}. A closed-loop control scheme was also implemented for Joint D, a simplified block diagram of which is shown in Fig.\,\ref{fig_control_system}. In the figure, (a) represents the input desired angle through time. Block (b) is the empirical tendon stroke-joint angle model. Block (c) is the low level motor control. Block (d) represents the physical robot, which is the plant of the system. Block (e) is a PID controller, with proportional, integral, and derivative constants of \SI{0.1}{} and \SI{0.1}{}, and \SI{0}, respectively. Block (f) represents the angle estimation procedure outlined in Section\,\ref{section_reconstruction}. The desired angle, actual angle, and estimated angle using the PI model are shown in Fig.\,\ref{fig_control}(b). The RMSE between these values are given in Table\,\ref{table_control_results}. The closed-loop control system reduced the tracking error between the actual and desired angle to \SI{2.92}{\degree} compared to \SI{8.14}{\degree}, demonstrating the usefulness of the proposed sensors for improving the accuracy of controlling the angle of the robotic joint. 

\subsection{Phantom Demonstration}\label{section_demo}

To demonstrate the sensing joint in a tissue-like medium, Joint D was used to navigate inside a hydrogel phantom. The phantom was made of \SI{7}{\percent} polyvinyl alcohol and \SI{0.85}{\percent} phytagel (both Sigma-Aldrich, Darmstadt, Germany) solution. A linear translation stage was added to the actuation system for Joint D to allow the advancement of the robotic joint into the phantom. An OEC 9800 Plus C-Arm system (GE Healthcare, Chicago, IL, USA) was used to take periodic (\SI{2}{\hertz}) fluoroscopic images of the joint to capture the ground truth deflection angle, since using the EM tracker in the hydrogel was not practical. A gold plated pin was affixed to the tip of the robot to increase its visibility under fluoroscopy. The acquired images were cropped, a binary threshold was applied to identify the shape of the pin, and an ellipse was fit to the shape of the pin. The angle between the ellipse's major axis and the horizontal in each image frame was taken to be the joint angle. The values were filtered with a moving average filter of width \SI{8}{}, and the mean value of the angle before bending commenced was subtracted from each angle value to obtain the change in angle. The joint was actuated in a pre-programmed manner while the sensor voltage was recorded. The sensor resistance was calculated and the joint angle estimated using the PI model and linear model obtained in Section\,\ref{section_control}. Three trials were conducted. The angles from the fluoroscopy images and the estimated angles from the sensor are shown in Fig.\,\ref{fig_phantom_demo}(a), Fig.\,\ref{fig_phantom_demo}(b), and Fig.\,\ref{fig_phantom_demo}(c) for Trial 1, Trial 2, and Trial 3, respectively. Selected fluoroscopy image frames before actuation, during bending, and near maximum bending for Trial 2 are shown in Fig.\,\ref{fig_phantom_demo}(d), Fig.\,\ref{fig_phantom_demo}(e), and Fig.\,\ref{fig_phantom_demo}(f), respectively. The RMSE of the PI and linear joint angle estimation for each trial are shown in Table\,\ref{table_demo_results}. Since the angle estimation from fluoroscopy and the DLW sensor matched closely, the sensor could potentially be used to supplement fluoroscopy guided navigation during a procedure to allow a lower frame rate and therefore decrease exposure for clinicians.

\section{Conclusion}\label{section_conclusion}
In this work, a method for fabricating DLW strain sensors directly on polymer notched-tube robotic joints was proposed and demonstrated. The fabricated sensing joints were characterized using both linear and nonlinear approaches. Rate dependence and relaxation were considered. The sensors were used to estimate the angle of the sensing joints, and it was shown that the sensor feedback could be used to implement closed-loop control. Finally, a demonstration of a sensing joint in a hydrogel phantom was presented. While the experiments showed low errors in estimating the bending angle and low tracking errors during closed-loop control, the sensing joints have opportunities for improvement. The sensing range is relatively low, with saturation observed at less than \SI{2}{\percent} strain. Additionally, the durability and temperature dependence of the sensors will need to be investigated further. Future work will also compare the performance of the proposed sensor to other shape sensing methods. Overall, the integration of DLW sensors directly onto continuum robotic joints represents a low-cost, easily fabricated method to obtain shape feedback. 

\bibliographystyle{IEEEtran}
\bibliography{references}

\end{document}